\documentclass[letterpaper, 10 pt, conference]{ieeeconf}  

\IEEEoverridecommandlockouts                              

\usepackage{graphics} 
\usepackage{epsfig} 
\usepackage{mathptmx} 
\usepackage{times} 
\usepackage{amsmath} 
\usepackage{amssymb}  

\usepackage{times}
\usepackage[font=footnotesize, skip=10pt]{caption}
\usepackage{multicol}
\usepackage[bookmarks=true]{hyperref}
\usepackage[utf8]{inputenc}
\usepackage{amsmath,amssymb}
\usepackage{algorithm}
\usepackage{algorithmic}
\usepackage{graphicx}
\usepackage{booktabs}
\usepackage{xcolor}
\usepackage{xspace}
\usepackage{verbatim}
\usepackage{siunitx} 
\usepackage{algorithm}
\usepackage{algorithmic}
\usepackage{float} 
\usepackage{cuted}  

\newcommand{\method}{SCOOP'D\xspace} 

\newcommand{\ba}{\mathbf{a}}

\newcommand{\bo}{\mathbf{o}}

\title{\LARGE \bf
\method: Learning Mixed-Liquid-Solid Scooping \\ via Sim2Real Generative Policy
}

\author{Kuanning Wang$^{1}$, Yongchong Gu$^{1}$, Yuqian Fu$^{3}$, Zeyu Shangguan$^{2}$, Sicheng He$^{2}$,\\ Xiangyang Xue$^{1}$, Yanwei Fu$^{1,\dagger}$, Daniel Seita$^{2}$
\thanks{
$^1$~Fudan University, China.
$^2$~University of Southern California, USA. 
$^3$~INSAIT, Sofia University “St. Kliment Ohridski”, Bulgaria. 
}
\thanks{$\dagger$ indicates the corresponding author}
}

\begin{document}

\maketitle
\thispagestyle{empty}
\pagestyle{empty}

\begin{strip}
\centering
\vspace{-50pt}
\includegraphics[width=\textwidth]{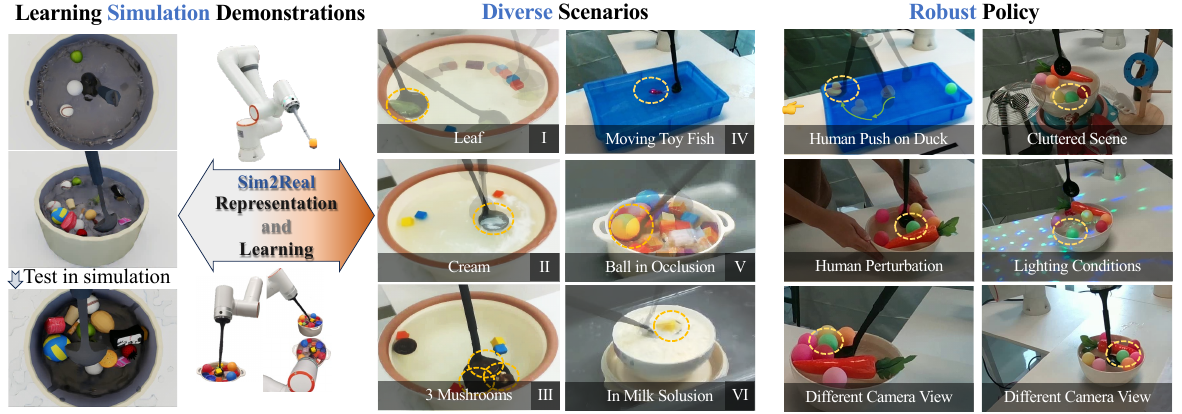}
\captionof{figure}{
  \textbf{Sim2Real Demonstrations and Generalization.}
  Our method, \method, is trained entirely in simulation (left) and generalizes to diverse real-world scenarios (middle) and robust conditions (right).
    The left column shows simulation demonstrations and testing examples.
    The middle column (I–VI) presents real-world scenes with various objects and environments, where yellow circles denote the target objects.
    The right column demonstrates the robustness of our learned policy under different disturbances such as human perturbations, lighting changes, and different camera viewpoints.
}
\label{fig:pull}
\end{strip}

\begin{abstract}

Scooping items with tools such as spoons and ladles is common in daily life, ranging from assistive feeding to retrieving items from environmental disaster sites. However, developing a general and autonomous robotic scooping policy is challenging
since it requires reasoning about complex tool-object interactions. Furthermore, scooping often involves manipulating deformable objects, such as granular media or liquids, which is challenging due to their infinite-dimensional configuration spaces and complex dynamics. 
We propose a method, \method, which uses simulation from OmniGibson (built on NVIDIA Omniverse) to collect scooping demonstrations using algorithmic procedures that rely on privileged state information. Then, we use generative policies via diffusion to imitate demonstrations from observational input. 
We directly apply the learned policy in diverse real-world scenarios, testing its performance on various item quantities, item characteristics, and container types.  
In zero-shot deployment, our method demonstrates promising results across 465 trials in diverse scenarios, including objects of different difficulty levels that we categorize as ``Level 1'' and ``Level 2.'' 
\method outperforms all baselines and ablations, suggesting that this is a promising approach to acquiring robotic scooping skills.
Project page is at https://scoopdiff.github.io/.
\end{abstract}


\section{Introduction}

Scooping with tools like ladles or spoons is a fundamental skill in tasks ranging from cooking~\cite{liu2022robotcookingstirfry}, assistive feeding~\cite{brose2010assistive,park2017multimodal} and environmental cleanup~\cite{ruangpayoongsak2017wastescooper}. While developing general-purpose robotic scooping could offer broad social and economic benefits, it remains a challenging problem due to the complexity of tool-object interactions and the need to handle mixtures of deformable materials such as liquids and granular media~\cite{manip_deformable_survey_2018,deformables_survey_2022}.
Also, observations are often unreliable due to occlusions and fluid surfaces that cause reflections, refractions, and unstable depth sensing, which makes scene representations noisy and unreliable.

Prior work has explored scooping a single item~\cite{Seita2022toolflownet} or using one type of granular medium~\cite{schenck2017learning}, which significantly simplifies the task. In contrast, we aim to advance scooping to the next level by addressing more realistic and challenging settings---scooping from containers that have a mixture of liquids and multiple solid items, where distractor objects are often present. To tackle this, we equip a robotic manipulator arm with a ladle as its end-effector and task it with retrieving a subset of target items from these complex environments.

Since collecting real-world data for robotic scooping is time-consuming, costly, and potentially dangerous, we use a \textbf{Sim2Real} learning paradigm for efficient, scalable, and safe data collection, as done in other robotics applications~\cite{openai-dactyl,domain_randomization}. 
Moreover, scooping floating objects requires smooth and precise control, but reinforcement learning suffers from low sample efficiency and jerky motions~\cite{Seita2022toolflownet}, motivating our imitation learning approach. 
Based on OmniGibson~\cite{li2022behavior}, a recent simulator built on NVIDIA Omniverse, we implement an algorithmic demonstrator that provides demonstrations while using ``privileged'' ground-truth object state information from simulation. We obtain a new simulated multimodal scooping dataset (\textbf{\textit {SimScoop}}) with 6,480 demonstrations.

We further propose a novel method,  \textbf{SCOOP}ing with \textbf{D}iffusion (\method), which learns robotic scooping from demonstrations (e.g., from \emph{SimScoop}). \method contains two Diffusion Policy models~\cite{DiffusionPolicy2023}, where one learns a good initial ``pre-scoop'' ladle pose and the other learns fine-grained scooping motions.
Our method is quick to train and benefits from advanced vision-based foundation models~\cite{Bommasani2021FoundationModels} such as SAM2~\cite{ravi2024sam2}.
Critically, as in Fig.~\ref{fig:pull}, our learned policy can be deployed directly to diverse real-world scenarios without any extra fine-tuning.
Quantitatively, our method achieves over 80\% success across 240 real-world trials on ``Level 1" objects, outperforming baselines and ablations. This is a challenging zero-shot setting under a strict success criterion: scooped objects must contain exactly the target(s) and no others. Extensive experiments show that \method generalizes across objects, multiple targets, varying occlusion severity (``Normal" and ``Severe"), liquids, and containers.

To summarize, the contributions of the paper include: 
\begin{itemize}
    \item A simulation-based environment in OmniGibson for synthesizing diverse scooping demonstrations, along with the resulting 6,480-demo \emph{SimScoop} data.
    \item The novel \method method for learning from state-based demonstrations in simulation for Sim2Real transfer that leverages generative policies via diffusion for pre-scoop pose estimation and scooping motions. 
    \item Extensive experiments across 465 real-world trials demonstrate promising and competitive results, with strong generalization across varied scenarios.
    
\end{itemize}

\section{Related Work}
\subsection{Scooping and Manipulation of Deformable Objects}
Scooping is a core challenge in robotic manipulation, with prior work addressing liquids alone~\cite{niu2023goatsgoalsampling}, liquid-solid mixtures~\cite{bhaskar2024lava}, granular media~\cite{schenck2017learning}, and dough~\cite{qi2024learninggeneralizabletooluseskills}. 
A common use case of robotic scooping is in assistive feeding~\cite{bimanual_food_2022,jenamani2024flair,park2017multimodal,sundaresan2023acquisition}, where a robot with a fork or spoon retrieves appropriately-sized food items from a plate or bowl to provide to a user. 
Our work takes inspiration from assistive feeding in designing a generalizable Sim2Real scooping pipeline to handle multiple (typically solid) objects in liquid. 
Other prior work in robotic liquid manipulation focuses on complementary tasks, such as pouring~\cite{narasimhan2022pouring,schenck2017visualclosedloopcontrolpouring,lin2023pourit} or understanding fluid dynamics~\cite{diff_fluid_dynamics_2018}.

Closely related prior work includes ToolFlowNet~\cite{Seita2022toolflownet}, SCONE~\cite{tai2023scone}, and LAVA~\cite{bhaskar2024lava}.
ToolFlowNet~\cite{Seita2022toolflownet} studies imitation learning from point cloud data and predicts dense 3D movement of tool points. During scooping, this approach outperformed reinforcement learning baselines which exhibited jerky and rapid motions. However, it assumes that the trajectory data is unimodal, but such data is often multimodal. 
It also struggles to track the object when displaced by the ladle.
SCONE~\cite{tai2023scone} uses active perception to interact with a solid or granular material in a bowl before scooping. 
LAVA~\cite{bhaskar2024lava} proposes a hierarchical policy framework that divides the task of 
scooping into high-level decision-making, mid-level action refinement, and low-level execution. 
Both SCONE and LAVA rely on demonstrations via kinesthetic teaching, which can be cumbersome to obtain. We use simulation to avoid collecting physical demonstrations.
We also study solid-\emph{liquid} manipulation, resulting in more complex object movements at test time compared to the tasks from~\cite{tai2023scone}.

\subsection{Imitation Learning from Simulation}
To learn scooping, we use imitation learning~\cite{imitation_survey_2018}, which trains a model to mimic actions from demonstrations. Imitation learning methods include behavioral cloning~\cite{pomerleau1989alvinn} and inverse reinforcement learning~\cite{ziebart2008irl}. More recent techniques predict a \emph{sequence} of actions to mitigate distribution shift~\cite{ALOHA}. In this direction, we leverage Diffusion Policy~\cite{DiffusionPolicy2023}, a popular approach for imitation learning that uses a diffusion model~\cite{ho2020ddpm} to effectively deal with multimodal and high-dimensional action distributions. 
In this work, we have two separate Diffusion Policy models. The first predicts a ``pre-scooping'' ladle pose for better initialization, while the second predicts fine-grained scooping actions.

To get demonstrations, we design an algorithmic demonstrator that uses ground-truth information in simulation. 
This strategy is inspired by other Sim2Real works such as fabric smoothing~\cite{seita_fabrics_2020} and scissor cutting~\cite{Lyu2024ScissorBotLG}, and we adapt it to scooping.
However, popular simulators in the robot learning community, such as PyBullet~\cite{coumans2019}, MuJoCo~\cite{mujoco}, and IsaacGym~\cite{makoviychuk2021isaac} do not support liquid manipulation.
Other works that study liquid manipulation in simulation include SoftGym~\cite{corl2020softgym}, FluidLab~\cite{xian2023fluidlab}, DAXBench~\cite{daxbench_2023}, and OmniGibson~\cite{li2022behavior}. We empirically find that OmniGibson has the best combination of simulation accuracy and usability.

\subsection{Perception for Sim2Real and Object Detection}
In this work, we learn a policy using Sim2Real, without real-world data collection. 
While techniques such as domain randomization over images~\cite{domain_randomization,cad2rl} have been beneficial for learning complex vision-based manipulation tasks~\cite{rubik_cube_2019,openai-dactyl}, there remains a large visual Sim2Real gap~\cite{reality_gap_1995} between scooping in simulation and the real world, where reflections and occlusions further hinder perception.
We address this gap by using a lower-dimensional state representation that consists of object poses.
Object pose estimation is a well-studied problem in robotics~\cite{PoseCNN} and in recent years, pre-trained foundation models~\cite{Bommasani2021FoundationModels} such as GroundingDINO~\cite{liu2024grounding,oquab2024dinov} and SAM2~\cite{ravi2024sam2} have facilitated generalizable pose estimation. We use these methods for real-time object segmentation and tracking, which helps us estimate object poses.

\section{Proposed Dataset and Method}
\noindent\textbf{Problem Statement:}
We study robotic scooping from a container on a tabletop with a mixture of liquid and solids. We use water as the liquid medium, which supports multiple floating solid objects. 
The robot is a standard manipulator equipped with a ladle as its end-effector. The robot executes actions to adjust the ladle's 6-DoF pose. 
An RGBD camera provides image data each time step. 
We define a \emph{trial} as an instance of the robot scooping task. At the start of each trial, a human places a mixture of items in the container. A text prompt informs the robot of the target item(s) to be scooped.

Since generalizable robotic scooping requires manipulating highly complex liquid-solid mixtures, we propose to learn scooping from demonstrations $\mathcal{D} = \{ \bo_1, \ba_1, \ldots, \bo_N, \ba_N \}$ consisting of observations $\bo_t$ and expert actions $\ba_t$ at each time step $t$. 
We use Diffusion Policy~\cite{DiffusionPolicy2023} and disentangle the scooping task into two key steps:
(i) reaching the pre-scoop ladle pose from a default pose and (ii) executing the subsequent trajectory. 
Consequently, we use two Diffusion Policy models for these steps. 
The first, $f_\phi$, predicts the pre-scoop ladle pose. The second, $\pi_\theta$, produces delta scooping actions $\ba_t$. 
To enhance the model's practicality, we equip it with the ability to determine the target item(s) based on the given text description. To achieve this, we first feed the text prompt to GroundingDINO~\cite{liu2024grounding,oquab2024dinov} to obtain the target bounding box(es). The box(es) are used as the visual prompt for SAM2~\cite{ravi2024sam2} which provides real-time object segmentation and tracking, enabling object pose estimation. 
Then, we design a geometry-aware network $g_\psi$ based on PointNet++~\cite{PointNet2_2017} to extract the object's states.

In the following, we present how we collect data in simulation and develop \method. See Fig.~\ref{fig:framework} for an overview.

\subsection{SimScoop Dataset}
\begin{figure*}[t]
  \centering
  \includegraphics[width=.95\linewidth]{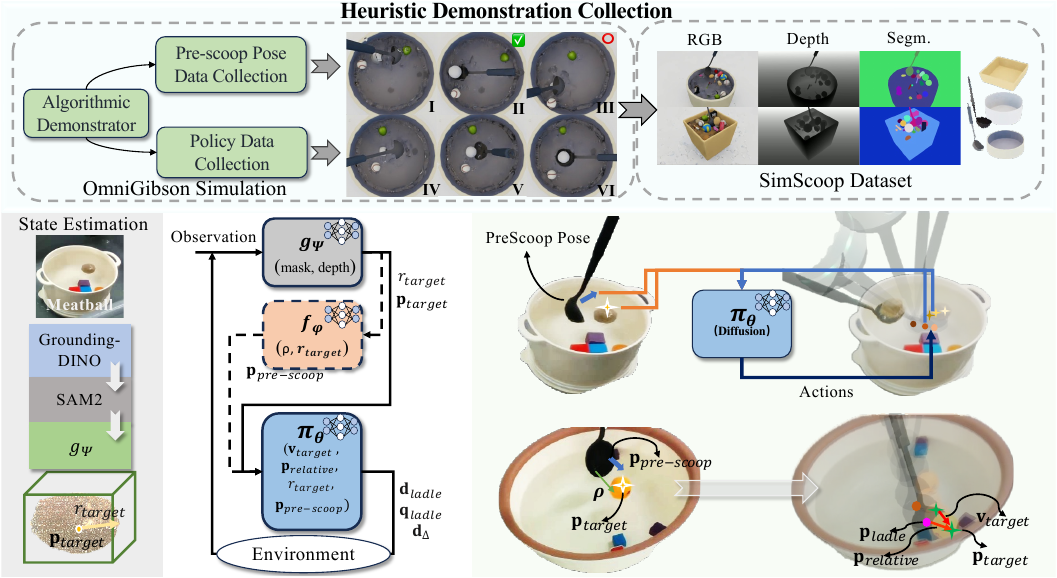}
  \vspace{-0.05in}
  \caption{
    \textbf{Our \method Method}. 
    The \underline{first} row shows the heuristic demonstration collection. Using OmniGibson simulation, we leverage an algorithmic demonstrator for SimScoop dataset collection. 
    The \underline{second} row shows how deployment works. The left part shows how we obtain the state of the target item from text (``meatball''), detection, live video stream segmentation, and regression with the partial point cloud. 
    The middle part shows the pipeline of our method. We use $f_\phi$ to generate a pre-scoop pose based on $\rho$ and $r_{\text{target}}$, then move the ladle directly to the generated pose. Then we leverage $\pi_\theta$ for closed-loop scooping. Our $\pi_\theta$ takes in $\mathbf{p}_\text{relative}$, $\mathbf{v}_\text{target}$, $\mathbf{p}_\text{pre-scoop}$ and $r_\text{target}$, and outputs $\ba_t$; $f_\phi$ is executed only once. 
    The right part shows the execution. We demonstrate the execution process in both the top and bottom containers, with the states specifically marked in the bottom for extra clarity.
    }
  \label{fig:framework} 
  \vspace{-0.08in}
\end{figure*}

\noindent\textbf{Simulation Environment.}
We use OmniGibson~\cite{li2022behavior}, which is powered by NVIDIA Omniverse. 
As shown in Fig.~\ref{fig:framework}, our simulation mainly contains different containers, several objects (e.g., balls), and ladles. The diameter and height of the first-row middle container are about \SI{0.4}{\meter} and \SI{0.2}{\meter}.
We modify the number of maximum micro-particle samples in the simulation environment to change the volume of water in the container.
However, we do not tune the physical parameters of the simulator to align it with real-world liquids.

\begin{figure}[t]
  \centering
  \includegraphics[width=.95\linewidth]{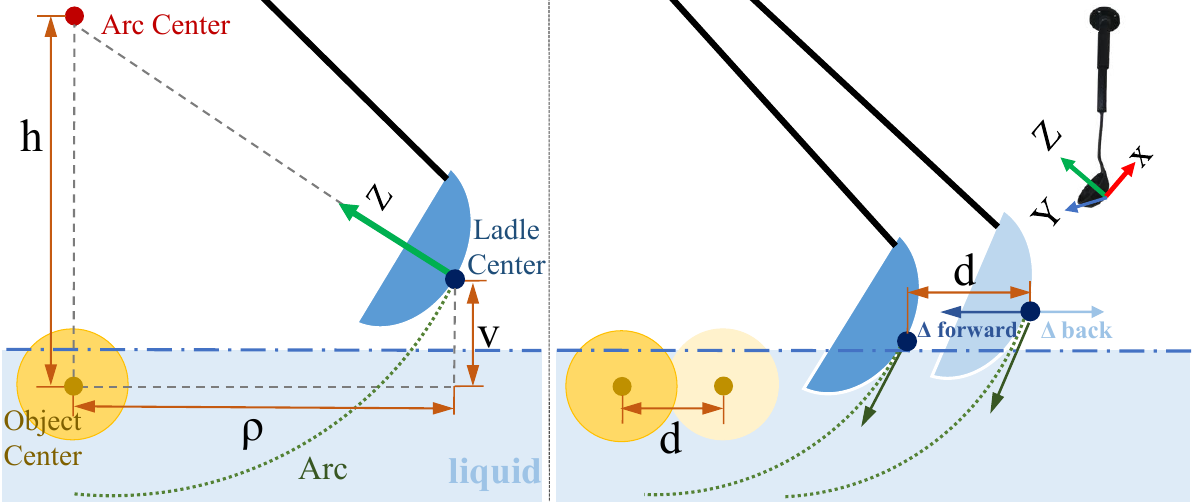}
  \vspace{-0.05in}
  \caption{
     \textbf{Heuristic scooping strategy.} We sketch a ladle and a target item (circle). The ladle's center of rotation is at the bottom of its ``bowl.'' It follows the dotted circular arc to go underneath the item, and then lifts up.
  }
  \label{fig:rule}
  \vspace{-0.15in}
\end{figure}

\label{ssec:rule_method}
\noindent\textbf{Heuristic Scooping Strategy.} To collect demonstrations, we implement a motion-adaptive heuristic scooping strategy that uses ground-truth simulation information; it defines a curve for the ladle's bowl to move it below the target and then lifts up the ladle. 
The curved trajectory allows gentler water entry, provides a gradual approach in cluttered scenes, and positions the ladle to enclose targets, which can help capture and improve adaptability to different object states. 
As sketched in Fig.~\ref{fig:rule}, the ladle's position is the origin of its coordinate frame, at the bottom center of its bowl. 
Let $v$ be the vertical distance from the target item (center) to the ladle, and $\rho$ be the horizontal distance from the target item to the ladle's bowl.
The ladle moves in a circular arc about a point which is at distance $h$ directly above the target. After this, the ladle's bowl is under the target and faces up.
Through these quantities, we can compute the ladle's pose and moving direction for each state. 
We directly use object position together with constraints among parameters to guide efficient sampling. 

However, this does not account for how the target may move, which requires shifting the ladle's trajectory.  
Let $d \in \mathbb{R}^3$ represent the small 3D movement of the target in a time step. 
To maintain a stable circular motion when the target moves, we first shift the ladle according to $d$, and then the ladle follows its circular trajectory. 
Furthermore, while this gives us a basic scooping trajectory, the ladle may contact the container wall while scooping.
If the ladle hits the wall, our heuristic strategy immediately moves the ladle away from it.
Once the ladle's bowl is sufficiently close and directly underneath the target, the ladle moves upwards to scoop it. 

\label{ssec:data_collection_sim}
\noindent\textbf{Collecting Simulated Training Data.}
We collect two datasets in simulation: a \textbf{small} dataset with 600 demonstrations and a \textbf{large} dataset with 6,480 demonstrations.

For the \textbf{small} dataset, we divide the simulated data collection into two parts: (i) a pre-scoop pose and (ii) a scooping motion. See Fig.~\ref{fig:framework} for a visualization. In both data collection phases, we use a ``PoolBall'' object as the target object, since it leads to relatively stable simulation performance. 
During our experiments, we generalize to scooping other objects. We also sample other objects as ``obstacles'' in simulation.

To collect ladle pre-scoop pose candidates, we sample object size and position, and sample ladle pose parameters to avoid object-ladle collisions. We initialize the ladle at this sampled pose and execute the heuristic scooping strategy. We retain a ladle pose for training only when the heuristic method achieves a successful scoop with a limited number of collisions before lifting.
This setup, inspired by pre-grasp poses~\cite{Dasari2022LearningDM}, provides coherent starting points for scooping, improving robustness, and facilitating efficient policy learning across varied trajectories. 
By explicitly generating valid pre-scoop poses, and training a generative model to predict poses (see Sec.~\ref{ssec:prescoop_pose_gen}) we assist the subsequent scooping.
For policy data collection, we also sample object size and position. The ladle is initialized by the learned pre-scoop pose generation module. We then apply the heuristic scooping strategy, inserting a “\textit{offset movement}” upon ladle-target collisions to slightly offset the ladle from the target (this step is omitted in pre-scoop pose collection). These interventions generate data for collision-recovery scenarios caused by imperfect initialization or movement noise.

When collecting training data, the pre-scoop pose collection has a 64\% success rate, while the policy demonstration has an 82\% success rate. We only keep successful cases.

As shown in Fig.~\ref{fig:framework}, we also construct more complex scenes with containers, ladles, and objects of varying types and sizes, along with more distractors.
We use the heuristic method to collect data in a single stage, by first sampling a pre-scoop pose and then executing a scoop. 
These setups introduce significant motion uncertainty due to ladle-object interactions, liquid flow, and clutter. 
The dataset includes RGB, depth, and segmentation images from three views, along with object and ladle states and motion data.
This \textbf{large} data has 6,480 demonstrations, each with 80 frames, totaling over 518k frames and 1.5M RGBD images.

\subsection{\method Methodology}
\noindent\textbf{Pre-Scoop Poses Generation.}
\label{ssec:prescoop_pose_gen}
We employ a pre-scoop pose, inspired by the human tendency to select a coherent entry point for smoothly scooping a target from the mixture. 
We use a 1D CNN-based Diffusion~\cite{DiffusionPolicy2023} model $f_\phi$, to generate pre-scoop ladle poses. 
The goal of $f_\phi$ is to produce a diverse set of effective pre-scooping poses for different containers, objects, and ladles, based on fundamental properties of the target object (such as its radius $r_{\text{target}}$) and human-controllable parameters ($\rho$).
The observations, which are the object's estimated radius $r_{\text{target}}$ and our manually specified value $\rho$, are fed into $f_\phi$, which predicts the values of $h$ and $v$. These predicted values, together with $\rho$ and the object's position, determine the ladle’s pre-scoop pose.

Meanwhile, the ladle starts from a random pose and directly moves to the pre-scoop pose, during which the object may move. Although the pre-scoop pose is fixed, the ladle tracks the object’s recent motion upon arrival, effectively achieving a state as if the object were stationary.

\noindent\textbf{Diffusion Policy for Scooping.}
We also use a 1D CNN-based Diffusion Policy model $\pi_\theta$ to execute scooping.
The \textbf{input} $\bo_t$ to $\pi_\theta$ is a 10D vector containing (i) the 3D relative position of the target object:
${\mathbf{p}_{\text{relative}} = \mathbf{p}_{\text{target}} - \mathbf{p}_{\text{ladle}}}$, 
(ii) the 1D estimated radius of the target $r_{\text{target}}$, (iii) the 3D parameterized representation of the pre-scooping pose 
${\mathbf{p}_{\text{pre-scoop}} = (\rho, h, v)}$, and (iv) the 3D movement of the target in the previous step:
${\mathbf{v}_{\text{target}} = \mathbf{p}_{\text{target}}^{(t)} - \mathbf{p}_{\text{target}}^{(t-1)}}$.

The \textbf{output} of $\pi_\theta$ is a 10D vector $\ba_t$ that contains (i) the 3D direction of ladle's main movement:
    ${\mathbf{d}_{\text{ladle}} = \left[ d_x, d_y, d_z \right]}$, executed with a constant step length $s$, (ii) the 4D ladle orientation represented by a unit quaternion ${\mathbf{q}_{\text{ladle}} = [q_x, q_y, q_z, q_w]}$, and (iii) 
    the 3D \textit{offset movement},
    ${\mathbf{d}_{\Delta} = \left[ \Delta x, \Delta y, \Delta z \right]}$.
These parameters determine the ladle's motion, 
while the ladle also directly follows $\mathbf{v}_{\text{target}}$ as mentioned in Sec.~\ref{ssec:rule_method}. 
Therefore, the aggregated motion of the ladle is: 
\[
\setlength{\abovedisplayskip}{-0.05pt}
\mathbf{d}_{\text{agg}} = s \cdot \mathbf{d}_{\text{ladle}} + \mathbf{d}_{\Delta} + \mathbf{v}_{\text{target}}.
\setlength{\belowdisplayskip}{-0.05pt} 
\]

As mentioned earlier, $\mathbf{p}_{\text{relative}}$ considers the relative motion between the object and the ladle. Our strategy aligns the object and ladle dynamics into a shared representation, which reduces the complexity and captures the essential motion patterns. 
Our $f_\phi$ outputs the pre-scoop pose, which is important for $\pi_\theta$ to understand the current state of the ladle and the heuristic trajectory to scoop the target object.
Therefore, we condition $\pi_\theta$ on parameters of the pre-scoop pose: $\rho$, $h$ and $v$. 
See the Appendix for model details.

\begin{figure*}[t]
  \centering
  \includegraphics[width=1.\linewidth]{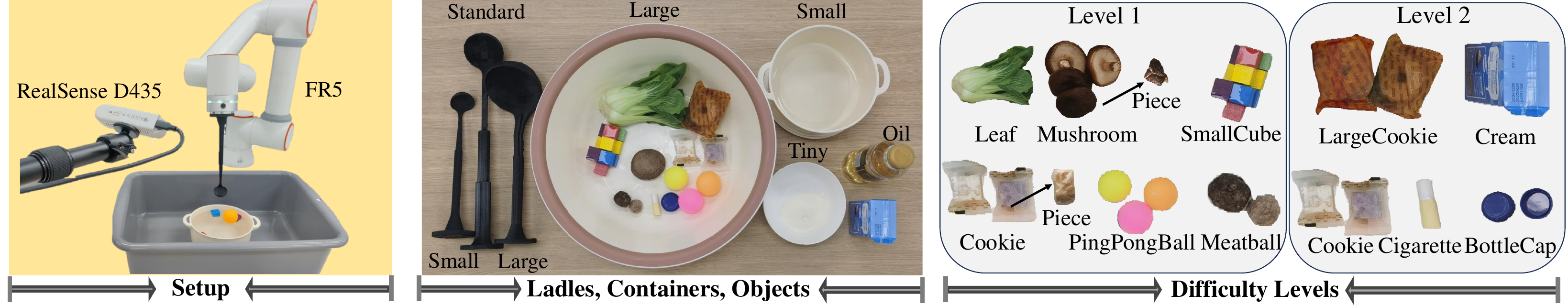}
  \vspace{-0.18in}
  \caption{
      \textbf{Real-world experimental setup.}
      \underline{\smash{Left}}: a third-person view of the setup. A third-person RealSense D435 camera captures RGBD image observations. 
      \underline{\smash{Middle}}: we show different ladles, containers, and objects that we use during physical experiments. The robot shown above (to the left) is holding the smallest ladle and operating on the small container (shown in the upper right corner). 
      \underline{\smash{Right}}: we show ``Level 1'' and ``Level 2'' (i.e., more challenging) objects that we use in our scooping experiments. See Sec.~\ref{ssec:exp_setup} for more details.
  }
  \label{fig:real_setup_objects}
  \vspace{-5pt}
\end{figure*}

\noindent\textbf{Geometry-Aware Localization and Scale Estimation.}
In addition to $f_\phi$ and $\pi_\theta$, our system incorporates a third learned component, a network $g_\psi$ based on PointNet++~\cite{PointNet2_2017}, which predicts an object’s center and longest radius from its point cloud. The longest radius serves as a scale reference that enables consistent handling across different shapes and helps reduce collision risks during manipulation.
We first use SAM2~\cite{ravi2024sam2} on the image to obtain a coarse segmentation of the target object. 
While SAM2 provides reasonable masks, directly estimating geometric properties from its segmentation can be unreliable under partial occlusions or segmentation noise, especially when liquids or surrounding obstacles limit depth visibility. 
Instead, we convert the segmented depth region into a point cloud and feed it into $g_\psi$, which captures fine-grained 3D geometric details to estimate the center and longest radius robustly.
To train $g_\psi$, we leverage the YCB dataset~\cite{calli2017ycbdata}, which offers diverse objects commonly used in manipulation tasks, including several basic shapes that resemble the custom objects used in our scooping scenario. 
Using PyRender~\cite{pyrender}, we render depth images of YCB objects, generate normalized point clouds, and train $g_\psi$ from scratch for accurate center and scale estimation.

\noindent\textbf{Sim2Real Transfer.}
To account for the Sim2Real physics gap, we add slight random noise to target and ladle motions during data collection for robustness. In deployment, we only calibrate the camera and deploy the policy directly.

\subsection{Application to Diverse Scooping Scenarios}

When we collect training data, we get scooping demonstrations of one object at a specific region of the container. However, we also apply \method to alternative scenarios.

\paragraph{Objects in Different Regions}
A practical difficulty with real-world robotic scooping is hitting the robot's kinematic limits. To address this for the circular containers we test, we divide each into four (non-equal) parts around the container’s center. The part directly opposite the robot and the two lateral parts are roughly equal in size, while the part closest to the robot is smaller and reflects where the robot may hit kinematic limits. 
We adjust the policy model’s input and output for consistency with demonstrations, applying region-based rotations: no rotation for the opposite area, 90° and 270° for the left and right, and 180° for the closest region. 
We do these straightforward calculations from standard observational data. 
Therefore, we can directly adapt the Diffusion Policy models $f_\phi$ and $\pi_\theta$ to handle targets everywhere in the container. 

\paragraph{Sequential Multi-Object Scooping}
Learning to scoop multiple objects directly from imitation learning is challenging, since a robot must prevent scooped objects from falling out, handle scattered object positions, and manage the size disparity between the ladle and objects. To address this, we sequentially scoop one item at a time, keeping previously scooped items in the ladle. After each scoop, the ladle lifts and moves to the pre-scoop pose for the next object.

\section{Experiments}
\label{sec:experiments}

\subsection{Experimental Setup}
\label{ssec:exp_setup}

\noindent\textbf{Simulation Setup.}
We construct two types of scooping environments by occlusion severity: (1) a simple setting with a single target object placed randomly in a large container (diameter \SI{0.4}{\meter}), and (2) a challenging setting with a smaller container densely filled with distractors, leading to severe congestion. In both settings, we use the same ladle tool and camera configuration (Fig.~\ref{fig:framework}, top center).
To simulate realistic conditions, the initial positions of all objects are randomized across trials.

\noindent\textbf{Real-World Setup.}
See Fig.~\ref{fig:real_setup_objects} for our setup. We use the low-cost (\$4,000) 6-DoF Fairino Robot 5 (FR5), and attach a ladle as its end-effector. 
We test three mixture containers with tiny, small, and large dimensions, and three ladles in small, standard, and large sizes, where size-matched ladles improve precision by avoiding obstacles in cluttered scenes. 
The experiments use objects inspired by assistive feeding and environmental clean-up applications. 
Example objects include food (e.g., mushrooms) and litter (e.g., bottle caps).

We categorize objects into two difficulty levels: Level 1 (easy) and Level 2 (hard).
Each object presents its challenges---such as the leaf being lightweight and easily moved, the PingPong ball’s fast motion, and the mushroom's irregular shape. 
See Figure~\ref{fig:real_setup_objects} for our categorization; the cookie can be either Level 1 or 2 depending on the ladle, as the size relationship between the ladle's bowl and the object significantly impacts the scooping difficulty.

As a pre-processing step before each trial, we move the robot to random positions within the environment and capture image observations. 
The pre-scoop pose is generated first, providing an ideal starting point, then the robot executes the policy $\pi_\theta$ to scoop the targets. 
We train $f_\phi$ on the small dataset, and $\pi_\theta$ on varying scales (small or large); unless noted, $\pi_\theta$ is trained on the small dataset.
See the Appendix for more details about the network and training process.

\begin{table*}[t]
  \centering
  \small %
    \scalebox{.85}{
  \begin{tabular}{c|c|c|c|c|c|c|c|c}
  \toprule
  \textbf{Occlusion Severity} & \textbf{Data Scale}  & Apple & PoolBall &  Strawberry & SoftBall &  Cork & Egg & Average \\
  \midrule
    Normal & Small (600 demos)  & 18/20 & 17/20 & 18/20 & 18/20 & 20/20 & 19/20 & 91.7\% \\
    Severe & Small (600 demos)  & 15/20 & 15/20 & 14/20 & 14/20 & 13/20 & 17/20 & 73.3\% \\
    Severe & Large (6,480 demos)  & 17/20 & 16/20 & 14/20 & 17/20 & 14/20 & 18/20  & 80.0\% \\
  \bottomrule
  \end{tabular}}
  \caption{ 
    \textbf{Scooping results in simulation.} We report the success rate of \method over 20 trials for each of six objects in simulation. 
    The last column averages the success across all six objects. All results are based on an action horizon of 1. 
    See Sec.~\ref{ssec:sim_results} for more details.
    }
  \label{tab:simulation_six_objs_occlusion_unify} 
  \vspace{-8pt}
\end{table*}

\begin{table*}
  \centering
  \small 
  \scalebox{.85}{
  \begin{tabular}{c|c|c|c|c|c|c|c} 
  \toprule
  Container-Ladle Size & PingPongBall & Mushroom & Cookie & SmallCube & Leaf & Meatball & Average \\
  \midrule 
   Large-Standard   & 18/20   & 17/20  & 17/20 & 16/20 & 16/20 & 17/20 & 84.2\% \\
   Small-Small      & 19/20   & 16/20 & 15/20 & 17/20 & 15/20  & 15/20    & 80.8\% \\
  \bottomrule
  \end{tabular}}
  \caption{  
    \textbf{Real-world scooping results.} We study the effect of the container and ladle size for scooping six items under ``Normal'' occlusion severity. We test with a large container and a standard ladle (top row) and a small container with small ladle (bottom row). We conduct 20 trials of \method and report the success rate. In all cases, we use an action horizon of 1. 
    We train $\pi_\theta$ on our small dataset.
    This table reports the \textbf{Level 1} objects for each ladle. 
  }
  \label{tab:real1} 
  \vspace{-16pt}
\end{table*}

\noindent\textbf{Evaluation Metrics.}
For single-object scooping, a success is scooping \emph{exactly} the target object above a height threshold. A failure is any other scenario, including when the target and non-targets are scooped together.
For multi-object scooping, we employ several evaluation metrics (see Table~\ref{tab:multiobj1}). 

\subsection{Baselines and Ablations}
\label{ssec:baselines_and_ablations}

\subsubsection{Baselines}
In the real world, we compare against six strong baselines:
\textbf{LAVA}~\cite{bhaskar2024lava}, \textbf{Heuristic Method} (Sec.~\ref{ssec:rule_method}),  \textbf{RGB-Based Diffusion Policy}, \textbf{Real-world State Diffusion Policy}, and the vision-language-action model from Physical Intelligence, \textbf{$\pi_0$ (zero-shot and fine-tune)}~\cite{Black20240AV}.

Since LAVA's code is unavailable, we re-implement its low-level wall-guided scooping policy for center and edge object placements. 
To simplify alignment, we use our segmentation module and manual object alignment, significantly reducing the difficulty---LAVA’s rule-based open-loop alignment struggles with floating objects. Items are manually placed near center or wall, with liquid-induced perturbations, and we evaluate success rates using its wall-guided policy.

For the heuristic method, since we test in real, we use estimated rather than ground-truth object states. 
Since detecting collisions from the camera is challenging, our adaptation omits the ``move-away'' action when the ladle collides with the container.
We test two versions of the method in real: one that uses our pre-scoop pose generation module, and another that relies on sampled pre-scoop poses, following the same process as pre-scoop pose data collection.

To compare with the following two baselines, we trained \method on 50 randomly sampled demonstrations from the original 600-demonstration dataset (in simulation).
For the RGB-based Diffusion Policy, we sampled 50 scooping demonstrations with ping pong balls and small cubes in the container (in the real-world), and train an end-to-end network. 
For the real-world state Diffusion Policy, we also use 50 manually demonstrated scooping trajectories with ping pong balls and small cubes. Object states of demonstrations were estimated via an external camera, and the policy was trained end-to-end without a pre-scoop pose. 

We also use $\pi_0$, a state-of-the-art vision-language-action model, as a baseline. The language instruction is one sentence, and requests to scoop the target. We test $\pi_0$ in two settings: zero-shot and fine-tuned on 50 demonstrations.

\subsubsection{Ablations}
We study four ablations trained on the original 600 demonstrations of \method: 
\underline{First}, we remove the pre-scoop pose generation $f_\phi$. Instead, we get the pre-scoop pose using the same sampling method as we did for collecting pre-scoop pose data. 
\underline{Second}, we randomly sample the pre-scoop poses within a sufficiently large sparse space, and then execute the policy. 
\underline{Third}, we remove the PointNet++ module $g_\psi$ by directly using the average position of the partially segmented point cloud as the object's center. The radius is determined by the longest distance from any point in the cloud to this center. 
\underline{Fourth}, for the target's relative motion, we directly input the target's position and the ladle's position into the policy, instead of using $\mathbf{p}_{\text{relative}}$.

\section{Results}
\label{sec:results}

\subsection{Simulation Results}
\label{ssec:sim_results}

Table~\ref{tab:simulation_six_objs_occlusion_unify} reports our main simulation results for scooping. We evaluate scooping six objects for 20 trials each, and for different occlusion severities and data scales for $\pi_\theta$. 
The success rate is 91.7\% under ``Normal'' occlusion severity and ``Small'' data scale. 
The table also presents results under ``Severe'' occlusions, where the success rate declines but is still relatively high at 73.3\%. During tests, we observe that the target frequently moves beneath the ladle and nearby objects. Training on our large dataset improves scooping performance in occluded scenes, increasing the success rate from 73.3\% to 80\%.
A “Severe” case is shown in the third row of Fig.~\ref{fig:meatball_scoop}.
We also observe two common failure modes. First, some ``correct'' scooping motions nonetheless fail when objects fall through the ladle bowl due to simulation inaccuracies in collision checks. Second, GroundingDINO~\cite{liu2024grounding} might not correctly detect objects, which may be because its training data is primarily based on real-world images.

\subsection{Real-World Results}
\label{ssec:real_world_results}

\begin{figure}[t]
  \centering
  \includegraphics[width=1.\linewidth]{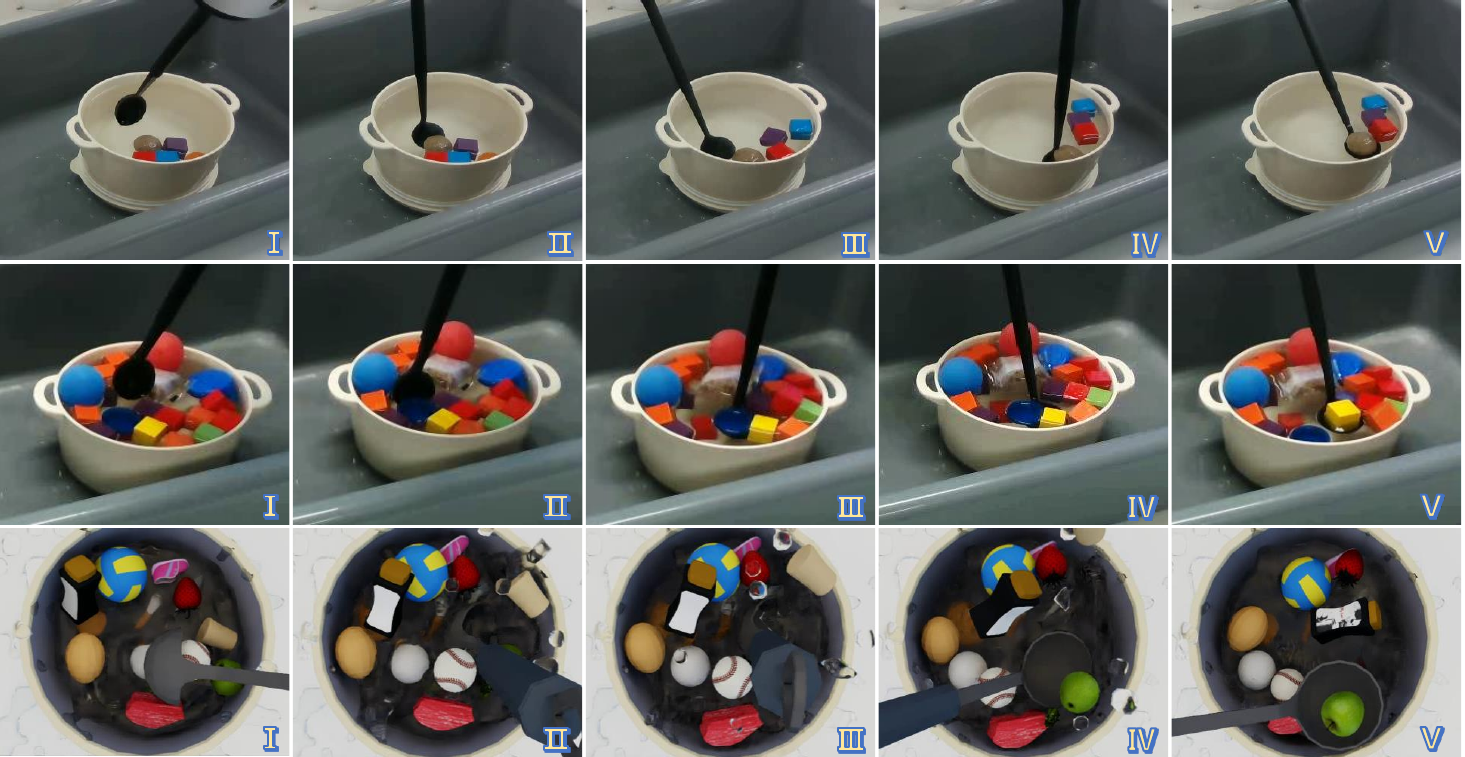}
  \vspace{-0.2in}
  \caption{
      \textbf{Visualization of Scooping.} We show our learned policy scooping targets in clutter. We show when it reaches the pre-scoop pose (I), when it moves towards the target (II, III, IV), and finally when it successfully scoops it (V).
      The first row shows the meatball in a mildly occluded scene, the second row depicts a yellow cube under heavy occlusion in real, and the third row presents a green apple in a severely occluded scene in simulation.
   }
  \label{fig:meatball_scoop}
  \vspace{-12pt}
\end{figure}

We show our main real-world scooping results in Table~\ref{tab:real1}.
\method achieves 82.5\% success rate across 240 trials with various Level 1 objects, containers, and ladles, highlighting the strong generalization ability of our Sim2Real approach.
Despite the challenging objects with diverse physical properties, the scooping performance among all objects is $\ge$75\% for each experiment setting. 
For different ladles, using the standard ladle shows a similar success rate as the small ladle, despite how we only collect the small dataset for one ladle size. 
Our method is also insensitive to the action horizon, as shown in the Appendix.

The failure cases mainly stem from kinematic limits of the robot hardware. One failure comes from losing sight of the object, which could be optimized by using multiple cameras or an eye-on-hand~\cite{hsu2022visionbasedmanipulatorsneedhands} camera. Other cases include where, while scooping the target object, other objects that we do not want to scoop are also lifted. 

See Fig.~\ref{fig:pull} (I) for a qualitative result. 
The leaf's irregular shape and lightweight nature makes it challenging to scoop, often causing it to float away. However, after several actions, the leaf is successfully scooped.
See Fig.~\ref{fig:meatball_scoop} for another example, where we scoop a meatball near obstacles using the small ladle. Unlike Fig.~\ref{fig:framework}, where the meatball is scooped out in one motion, obstacles affect the perception of its state, causing it to slide away. 
Both of the qualitative results indicate that \method generates valid pre-scoop poses, and that our policy can scoop in challenging dynamic situations.

\paragraph{Scooping Multiple Objects}
See results in Table~\ref{tab:multiobj1} for scooping up to 3 objects. 
For PingPongBall scooping, the average number of scooped targets is higher when scooping 2 targets compared to 3. The main reason is that when scooping a new object, the ladle needs to hold the previously scooped items. As a result, there are often failures when scooping the third item because the ladle cannot contain the first two.
On the other hand, for objects like mushrooms with irregular shapes or small cubes, it is easier to keep the scooped items stable without them falling off the ladle or flowing away while scooping additional items.
The probability of obstacles being scooped increases as more objects are scooped.
See Fig.~\ref{fig:pull} (III) for a qualitative result. 

\paragraph{Different Liquids}
We test changing the liquid and observe that \method shows some degree of ``liquid generalization.'' Due to space limitations, we defer details to the Appendix. See Fig.~\ref{fig:pull} (VI) for a qualitative result.

\begin{table}
  \centering
  \small 
   \scalebox{.95}{
  \resizebox{\columnwidth}{!}{%
  \begin{tabular}{c|c|c|c|c} 
  \toprule
  \textbf{Object (Count)} & average & average & success & success \\
                          & targets ($\uparrow$) & obs ($\downarrow$) & w/o obs ($\uparrow$) & w/obs ($\uparrow$) \\
  \midrule
  SmallCube (1)       & 1   & 0   & 5/5 & 5/5 \\
  SmallCube (2)       & 1.8 & 0   & 4/5 & 4/5 \\
  SmallCube (3)       & 2.4 & 0.4 & 3/5 & 1/5 \\
  PingPongBall (1)    & 0.8 & 0.2 & 4/5 & 3/5 \\
  PingPongBall (2)    & 1.6 & 0.6 & 4/5 & 1/5 \\
  PingPongBall (3)    & 1.2 & 0.2 & 1/5 & 1/5 \\
  Mushroom (1)        & 0.8 & 0   & 4/5 & 4/5 \\
  Mushroom (2)        & 1.6 & 0.2 & 4/5 & 3/5 \\
  Mushroom (3)        & 1.6 & 0   & 1/5 & 1/5 \\
  \bottomrule
\end{tabular}}
  }
  \caption{
    \textbf{\method results for scooping multiple objects in the real world.}
    SmallCube: For 1 and 2 objects, we use the standard ladle size.
    For 3, we use the large version.
    PingPongBall: We use the large ladle to scoop 1, 2 and 3 balls, as it can hold up to three PingPongBalls.
    Mushroom: We use the large ladle to scoop 1, 2, or 3 mushrooms. This ladle can also hold up to three. 
    For each row, we present (respectively): the average number of scooped targets (higher is better), the average number of obstacles scooped (lower is better), the success rate of scooping a specific number of targets without and then with considering obstacles (w/o obs, w/obs). 
  }
  \label{tab:multiobj1} 
  \vspace{-5pt}
\end{table}


\begin{table}
  \centering
  \small %
  \scalebox{.8}{
  \begin{tabular}{c|c|c|c|c} 
  \toprule
  \textbf{Occlusion}  & \textbf{Data Scale} &  SmallCube & PingPongBall  & Average \\
  \midrule
    Normal & Small (600 demos)  &  17/20  & 19/20    &    90.0\%           \\
    Severe & Small (600 demos)  &  12/20  & 12/20    &   60.0\%           \\
    Severe  & Large (6,480 demos)  &  15/20  & 14/20    &    72.5\%           \\
  \bottomrule
  \end{tabular}}
  \caption{ 
  \textbf{Comparisons of different objects and occlusion severity for real-world scooping.} See Sec.~\ref{ssec:real_world_results} for more discussion. The ``Normal'' results are the same as those of SmallCube and PingPongBall in Table~\ref{tab:real1}.
  }
  \vspace{-15pt}
  \label{tab:real_occlusion} 
\end{table}

\paragraph{Different Occlusion Severity}
In Table~\ref{tab:real_occlusion}, we show experimental results with varying numbers of distractors. 
The ``Normal'' setting corresponds to the first row in Fig.~\ref{fig:meatball_scoop} (distractors \textless30\% of surface), while the ``Severe'' setting (see the second row of Fig.~\ref{fig:meatball_scoop}) involves heavy occlusion with a dense container filled with distractors. 
We compare the scooping performance for different data scales and occlusion levels. Compared to the ``Normal'' severity, performance drops under ``Severe'' occlusion, as expected. However, training on our larger dataset increases the performance under the same ``Severe'' occlusion, demonstrating the potential for improvements by scaling up the data.

\paragraph{Data Scalability}
We test scalability by training on datasets of different sizes. As shown in Table~\ref{tab:simulation_six_objs_occlusion_unify} and~\ref{tab:real_occlusion}, success rates improve in occluded scenes, indicating that our model effectively benefits from larger and more diverse data.

\begin{table}
  \centering
  \small %
\scalebox{.8}{
  \begin{tabular}{c|c|c} 
  \toprule
  \textbf{Method}  & SmallCube & PingPongBall   \\
  \midrule
    \method (ours, 600 demos)  &  17/20  & 19/20                         \\
    \method (ours, 50 demos)  &  13/20  & 14/20            \\
    \midrule
    \multicolumn{3}{c}{\textbf{Baselines}} \\
    RGB-Based Diffusion Policy (50 demos)  & 4/20  &   3/20       \\
    Real-world State Diffusion Policy (50 demos) & 6/20 &  5/20      \\
    $\pi_0$ (zero-shot)~\cite{Black20240AV} (50 demos) &   0/20    &   0/20 \\ 
    $\pi_0$ (fine-tune)~\cite{Black20240AV} (50 demos) &  2/20   &   4/20 \\ 
    LAVA*~\cite{bhaskar2024lava} &  7/20  & 6/20     \\
    Heuristic, sampled pre-scoop pose & 5/20   & 16/20  \\
    Heuristic, w/pre-scoop pose & 9/20  & 16/20 \\
    \midrule
    \multicolumn{3}{c}{\textbf{Ablation Studies}} \\
    \method sampled pre-scoop pose (600 demos)  &   15/20    &     18/20   \\ 
    \method w/o pre-scoop pose (600 demos)   &   0/20    &     0/20   \\
    \method w/o PointNet++ (600 demos)     &    14/20       &     17/20   \\
    \method w/o relative motion (600 demos)  &   7/20        &    14/20      \\
  \bottomrule
  \end{tabular}}
  \caption{ 
    \textbf{Comparisons of different methods and ablations for real-world scooping.} 
    The ``*'' denotes simplified conditions for LAVA. 
    See Sec.~\ref{ssec:baselines_and_ablations} and~\ref{ssec:real_world_results} for more discussion.
    The \method results are the same as those of SmallCube and PingPongBall in Table~\ref{tab:real1}.
  }
  \vspace{-0.2in}
  \label{tab:baseline} 
\end{table}

\paragraph{Baselines}
For LAVA, results in Table~\ref{tab:baseline} show that \method outperforms LAVA; we observe that LAVA's fixed trajectories do not scale well. A slight disturbance can cause objects to move away, making them difficult to reach the center. 
Moreover, pushing floating objects to the wall and scooping can also cause drift or squeeze them out.

For the ``Heuristic, sample pre-scoop pose" method, when scooping the small cube, the ladle often collides with the cube before reaching underneath it. This causes the cube or even the container to move heavily, making it even harder to scoop using the heuristic method compared to our policy.

For ``Heuristic, w/pre-scoop pose," the success rate of scooping the small cube improves as our $f_\phi$ generates more reliable poses. However, without the policy to adjust the movement, collisions still occur. Additionally, the heuristic methods are rigid and do not generalize well in the real world, leading to collisions during the ladle's movement.

For the ``RGB-Based Diffusion Policy,'' visual features fail to accurately capture the spatial relationship between the ladle (underwater) and the object, leading to imprecise scooping. The ladle often misses the target, and even when it gets close, suboptimal contact angles or positions tend to push the object out of the container, causing failure. 

For ``Real-world State Diffusion Policy'', 
unlike in simulation, where a structured approach first moves the ladle to a pre-scoop pose, the real-world policy directly predicts robot motions in a larger action space, increasing reliance on abundant high-quality data and reducing training efficiency.

For ``$\pi_0$ (zero-shot),'' the robot moves almost randomly. For ``$\pi_0$ (fine-tune),'' it moves toward the target but inaccurately, often colliding and failing. This may be because $\pi_0$ is mainly trained on tasks with grippers, and thus lacks exposure to ladle use and scooping applications.

\paragraph{Ablations}
We show results in Table~\ref{tab:baseline}. 
The ``sampled pre-scoop pose"  shows a slight decrease in the success rate for the small cube. Qualitatively, there are significantly more collisions when the ladle gets closer to the cube, causing the ladle to push the object and occasionally collide with the container. Despite this, due to our well-trained policy, the cube can still be scooped after a few additional steps.
For ``w/o pre-scoop pose," 
while our trained policy can often get closer to the target, accurately scooping the object in a closed-loop manner is still highly challenging.
For ``w/o PointNet++," we attain close (but worse) performance compared to \method, suggesting that our PointNet++ module is necessary for accurate object state predictions. Failure cases here are largely due to inaccurate object center estimations.
For ``w/o relative motion," the success rate is much lower than that of \method. Intuitively, relative motion reduces the complexity, making optimization easier.

\begin{table}
  \centering
  \small %
  \scalebox{.76}{
  \begin{tabular}{c|c|c|c|c|c|c} 
  \toprule
  \textbf{}  & LargeCookie & Cookie & BottleCap & Cigarette & Cream & Average \\
  \midrule
    \method       &11/20  &  9/20  &  9/20      &  12/20     &  7/20 & 48.0\% \\
    
  \bottomrule
  \end{tabular}
  }
  \vspace{-5pt}
  \caption{ 
    \textbf{Real-world \method results on Level 2 items.
    } 
  }
  \label{tab:level2} 
  \vspace{-18pt}
\end{table}

\paragraph{Scooping Level 2 Objects}
Table~\ref{tab:level2} reports results for scooping challenging objects with \method, 
which are lower than in Table~\ref{tab:real1} but still show robustness under these highly challenging unseen conditions.
For LargeCookie, we use a standard-size ladle to scoop the brown large cookie with wrapping paper, and for Cookie, we use the small ladle to scoop the purple cookie with wrapping paper. This is challenging since the objects have an uneven mass distribution and are much larger than the ladle's bowl.  
Other objects bring their own challenges: BottleCap can sink, SAM2 often fails to segment Cigarette, and Cream spreads easily.

\textbf{Analysis on limitations.}
While \method shows some generalization, it may struggle with more complex scooping tasks, such as scooping submerged objects deep in the container or deformable items such as cream.
Moreover, \method takes pretrained GroundingDINO and SAM2 models directly for object localization, without finetuning them for better performance.
Addressing these failures could be topics for future work.

\section{Conclusion}

In this paper, we present \method, a method for learning scooping policies from efficient algorithmic demonstrators in simulation. Our method uses Sim2Real generative models to imitate scooping behavior. Experimental results across a variety of real-world scooping scenarios suggest that \method obtains promising success rates. 
We hope this inspires future work on autonomous and generalizable robotic scooping.



\clearpage
\section{Appendix}

In the Appendix, we first provide additional implementation details of \method in App.~\ref{app:additional_method}. Next, we present extended experimental information in App.~\ref{app:additional_experiments}, followed by qualitative results demonstrating the generalization and robustness of our policy in App.~\ref{app:general_robustness}. Finally, we include further discussion of our work in App.~\ref{app:more_discussion}.

\subsection{\textbf{Additional Details of \method}}
\label{app:additional_method}

\subsubsection{Model and Training Details}

We collect 150 data points to train our pre-scoop pose generation model. 
We also collect two datasets in simulation: a small dataset with 600 demonstrations and a large dataset with 6,480 demonstrations to train the policy $\pi_\theta$.
When collecting data, we randomize the ladle's direction and orientation to make our learned models robust to these variations. 

We train our pre-scoop pose generation model for 1,000 epochs on a single NVIDIA A6000 GPU.
Similarly, we train Diffusion Policy models $\pi_\theta$ until the error stabilizes.
Both $f_\phi$ and $\pi_\theta$ use a 1D CNN-based~\cite{DiffusionPolicy2023} Diffusion Policy. The observation includes 2 steps, and the output from the current state to the future includes 3 steps for the action horizon.

\begin{table*}[t]
  \centering
  \small %
  \scalebox{.95}{
  \begin{tabular}{c|c|c|c|c|c|c|c}
  \toprule
   \textbf{Action Horizon} & Apple & PoolBall &  Strawberry & SoftBall &  Cork & Egg & Average \\
  \midrule
   1 & 18/20 & 17/20 & 18/20 & 18/20 & 20/20 & 19/20 & 91.7\% \\
   2 & 19/20 & 17/20 & 17/20 & 18/20 & 19/20 & 19/20 & 90.8\% \\
   3 & 19/20 & 20/20 & 16/20 & 18/20 & 17/20 & 18/20 & 90.0\% \\
  \bottomrule
  \end{tabular}
  }
  \caption{ 
    \textbf{Scooping Results in Simulation under Different Action Horizon.} We report the success rate of \method over 20 trials for each of six objects in simulation. The action horizon represents the prediction horizon for $\pi_\theta$, and the last column averages the success across all six objects. The \textbf{Action Horizon} 1 results are the same as those in Table~\ref{tab:simulation_six_objs_occlusion_unify}. All experiments are conducted under normal occlusion severity using our small dataset of 600 demonstrations.
  }
  \label{tab:simulation_six_objs_action} 
  \vspace{-0.05in}
\end{table*}

\begin{table}
  \centering
  \small 
  \begin{tabular}{c|c|c} 
  \toprule
  \textbf{Action Horizon} & PingPongBall & Mushroom  \\
  \midrule
    1 & 18/20  & 17/20 \\ 
    2 & 18/10  & 18/10 \\
    3 & 16/10  & 15/10 \\
  \bottomrule
  \end{tabular}
  \caption{ 
    \textbf{Scooping Results in Real under Different Action Horizon.}
    We use the PingPongBall and Mushroom objects for testing \method in the real world, where $\pi_\theta$ predicts 1, 2, or 3 action steps.
    The \textbf{Action Horizon} 1 results are the same as those of PingPongBall and Mushroom in Table~\ref{tab:real1}.
  }
  \label{tab:realacthor} 
  \vspace{-5pt}
\end{table}

\begin{table}[t]
  \centering
  \small 
  \begin{tabular}{c|c|c|c} 
  \toprule
  \textbf{ContainerSize}  & Water & PlantOil & MilkPowder  \\
  \midrule
    Smaller & 5/5 & 5/5 & 5/5 \\
  \bottomrule
  \end{tabular}
  \caption{ 
    \textbf{Experiment on different liquid scooping in the real world.}
    We test scooping a small yellow cube with our small ladle in the tiny container shown in Fig.~\ref{fig:real_setup_objects}. 
  }
  \label{tab:liquids} 
\end{table}

\subsubsection{Visualization of Pre-Scoop Poses}

\begin{figure}[t]
  \centering
  \includegraphics[width=1.\linewidth]{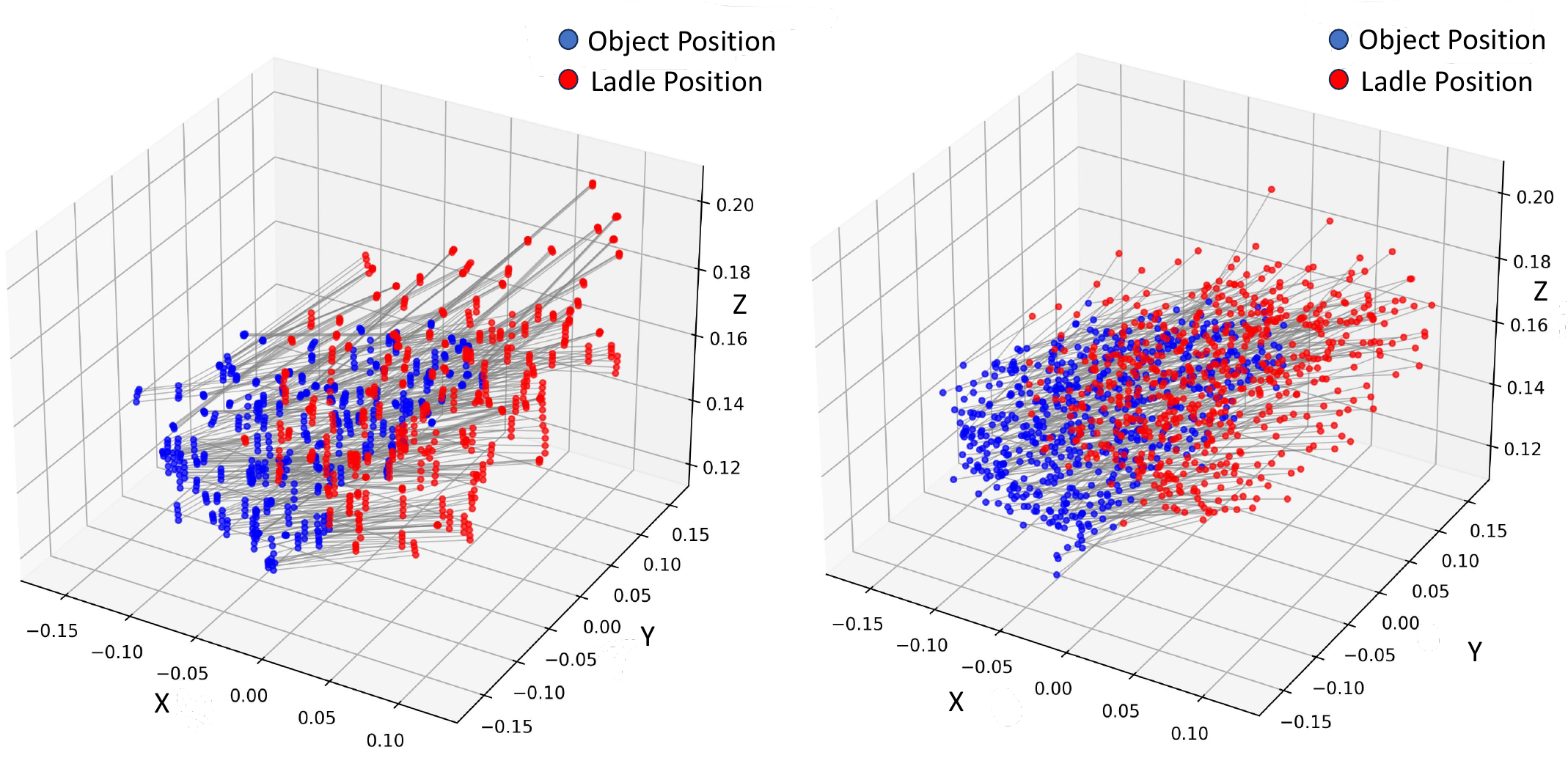}
  \vspace{-0.1in}
  \caption{
      \textbf{Visualization of Pre-Scoop Poses During Demonstration Collection and Inference.}
      We show the pre-scoop demonstration collection (left) and the inference process during policy demonstration collection (right).
      We plot the 3D positions of the ladle (red points) and the item from the demonstration data (blue points). 
      The blue points are randomly sampled, while the red points on the left are sampled, and those on the right are generated.
      We use gray lines to connect corresponding points.
  }
  \label{fig:prescoop_vis}
\end{figure}

Fig.~\ref{fig:prescoop_vis} provides a detailed view of the pre-scoop poses, where we visualize both the collected pre-scoop poses and the inferred pre-scoop poses from the policy demonstrations. As shown in Fig.~\ref{fig:prescoop_vis}, the collected pre-scoop poses exhibit significant variation, and our generation module, $f_\phi$, learns a similarly diverse distribution conditioned on the object's radius and $\rho$. 
The data collected in the left sub-figure contains positions where objects are sparsely distributed, while the policy data collected in the right sub-figure is denser and distributed uniformly.
Additionally, the right sub-figure illustrates that the pre-scoop poses used for policy demonstrations are diverse, enabling policy $\pi_\theta$ to learn from a wide range of conditions. 
This enhances the policy's robustness across different pre-scoop configurations and object types, ultimately facilitating the integration of both policies.

\subsubsection{Time for Data Collection and Training}

The time for collecting pre-scoop pose data is about half an hour, and the time for collecting policy data is about two hours (600 demos). The total training time for $f_\phi$ and $\pi_\theta$ is about one hour using a single NVIDIA A6000 GPU.

\subsection{\textbf{Additional Experiments}}
\label{app:additional_experiments}

In Table~\ref{tab:liquids}, we show the result of \method while scooping different liquids. The default in this paper is Water, and we test PlantOil and MilkPowder. We observe that we get 5/5 scooping performance under these particular settings. In future work, we will investigate liquids with substantially different viscosity properties.

Table~\ref{tab:simulation_six_objs_action} reports our scooping results in simulation, where we test scooping six objects for 20 trials each, and for 3 values of the action horizon for $\pi_\theta$. 
The results indicate that, when averaging over the action horizon, the success rates for \method are at least 90\% for all six objects. Furthermore, as shown in Table~\ref{tab:realacthor}, our real-world results are generally consistent across different action horizons.

\begin{figure}[t]
  \centering
  \includegraphics[width=1.\linewidth]{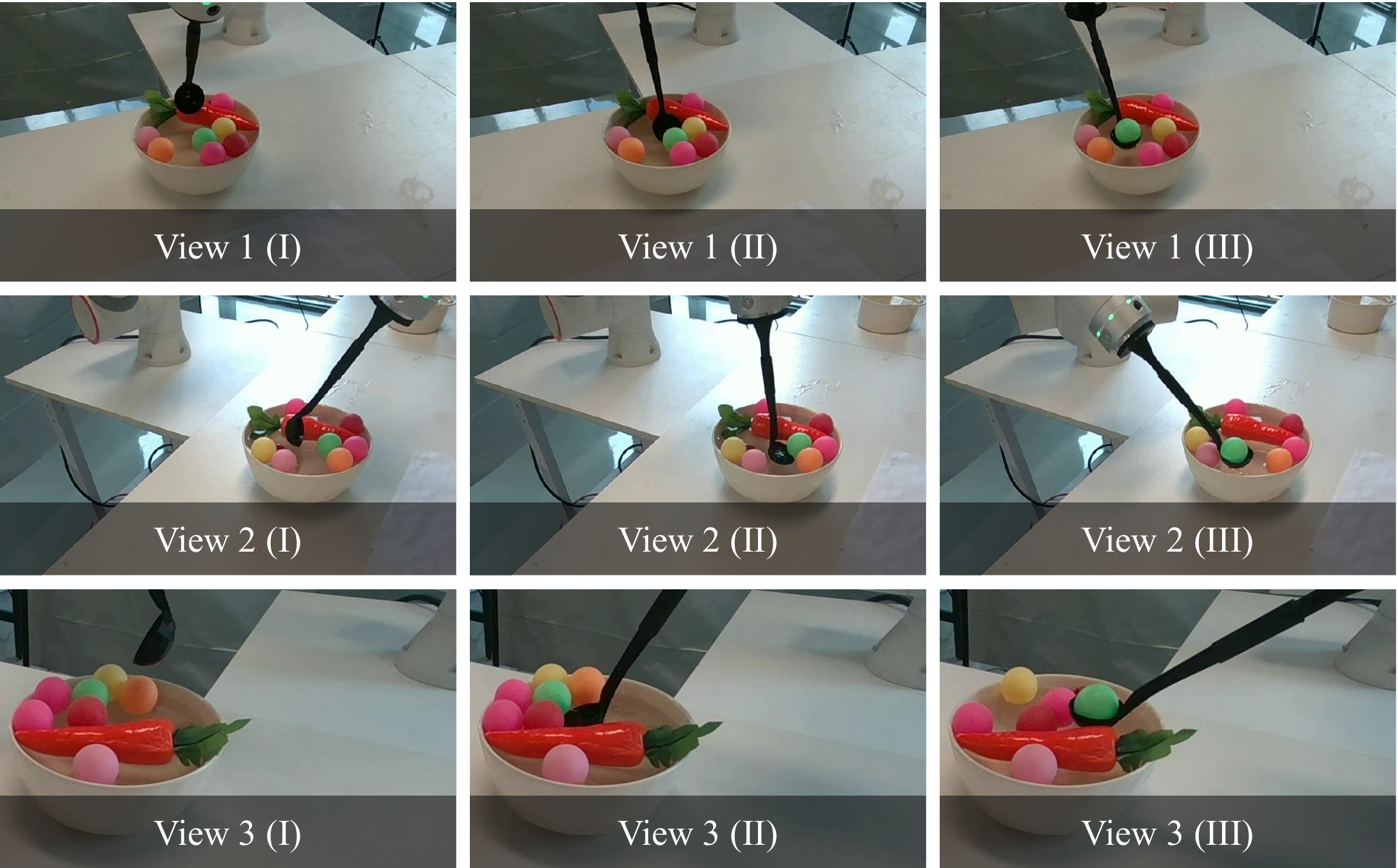}
  \vspace{-0.1in}
  \caption{
      \textbf{Visualization of Scooping in Different Viewpoints.} Each row illustrates the scooping process observed from a distinct camera viewpoint, where the robot attempts to scoop the green PingPongBall. Our policy performs consistently well across various camera views without any finetuning.
  }
  \label{fig:view_gen}
\end{figure}

\begin{figure*}[t]
  \centering
  \includegraphics[width=1.\linewidth]{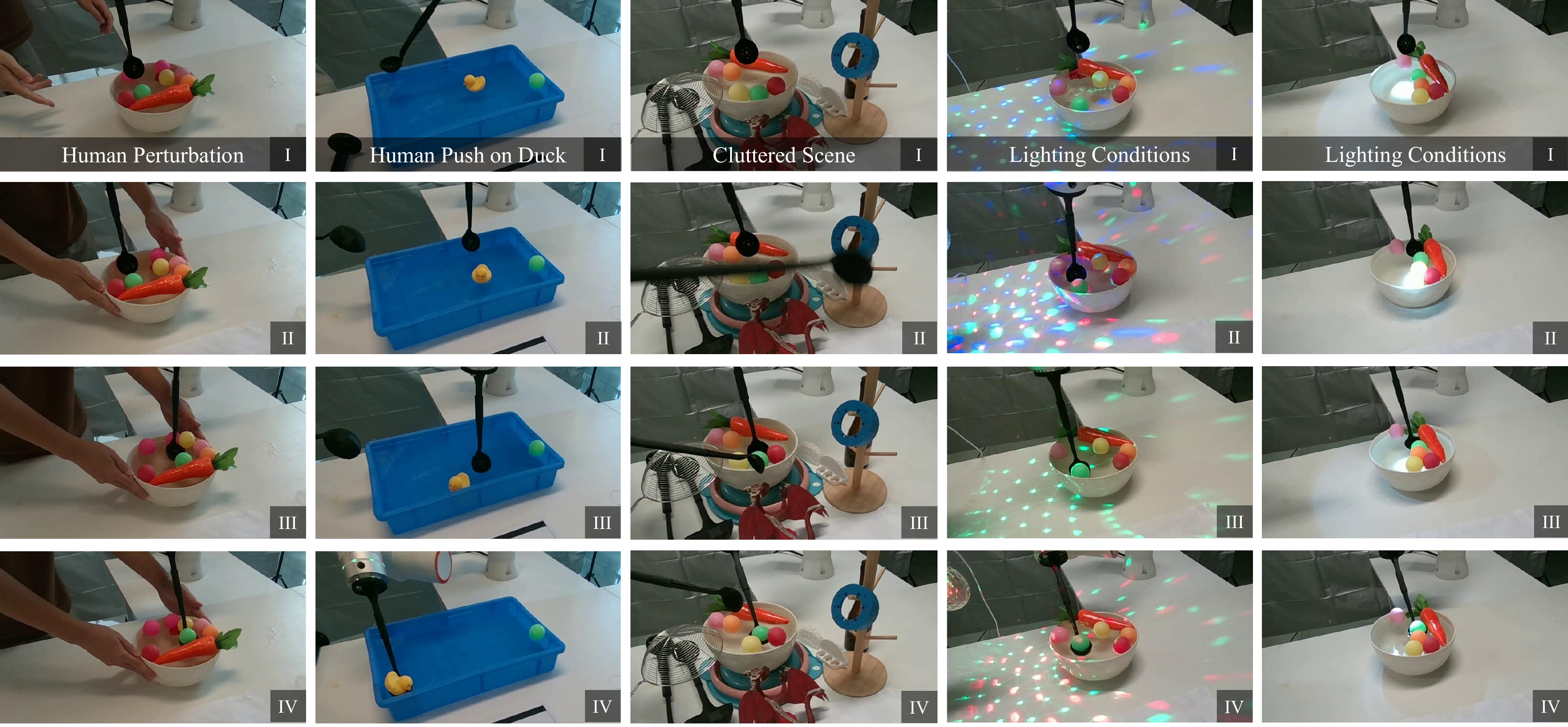}
  \vspace{-0.15in}
  \caption{
      \textbf{Visualization of Robustness Test.} Each column presents one type of external disturbance, under which our policy consistently performs well and demonstrates strong robustness.
  }
  \label{fig:robustness}
\end{figure*}

\begin{figure*}[t]
  \centering
  \includegraphics[width=1.\linewidth]{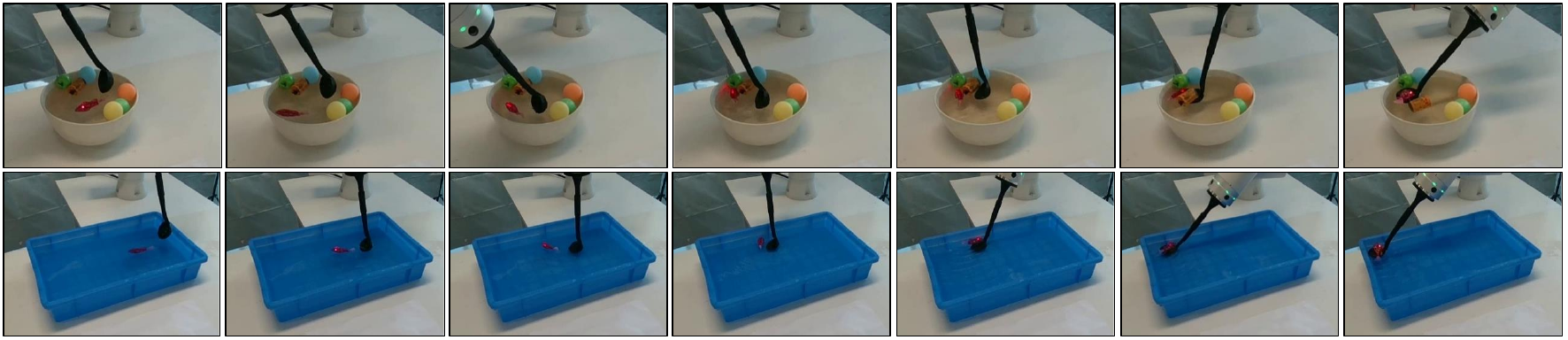}
  \vspace{-0.15in}
  \caption{
      \textbf{Visualization of Moving Toy Fish Scooping.} Each row illustrates one complete scooping attempt, shown as a sequence of frames. The robot attempts to scoop a self-moving electronic toy fish in these containers.
  }
  \label{fig:moving}
\end{figure*}

\subsection{\textbf{Generalization and Robustness Tests}}
\label{app:general_robustness}

\subsubsection{Viewpoint Generalization}

Our method can be directly deployed in the real world after camera calibration, without any additional training. All our real-world experiments were conducted from viewpoints different from those used in simulation. As shown in Fig.~\ref{fig:view_gen}, the qualitative results demonstrate that the policy robustly scoops the target green object under different camera viewpoints, indicating strong viewpoint generalization of our method.

\subsubsection{Robustness}
As shown in Fig.~\ref{fig:robustness}, our method demonstrates strong robustness under various challenging real-world conditions. 
Specifically, we test five types of perturbations: (1) \textbf{Human Perturbation}, where a person continuously moves the container to test dynamic scene adaptation; (2) \textbf{Human Push on Duck}, where a toy duck is intermittently pushed to different positions to evaluate the policy’s stability under external interference and moving targets; (3) \textbf{Cluttered Scene}, where both the inside and surroundings of the container are highly cluttered, and occasional human occlusions further challenge visual perception; (4) \textbf{Lighting Conditions (Colored Lights)}, where lights of various colors flash continuously to test robustness to changing illumination; and (5) \textbf{Lighting Conditions (White Light Reflection)}, where a flashlight is directed at different positions on the liquid surface to examine resilience against strong white light and surface reflections.
It consistently performs well despite external disturbances, significant changes in lighting, reflections on the water surface, and cluttered or visually complex backgrounds. These results highlight the stability and adaptability of our policy when deployed in diverse and uncontrolled environments.

\subsection{\textbf{Further Discussion}}
\label{app:more_discussion}

Although \method focuses on scooping solids from liquid, we also provide qualitative results in Fig.~\ref{fig:moving} showing that \method can scoop movable toy fish.
Besides, \method demonstrates strong robustness to various perturbations in Fig.~\ref{fig:robustness}, view generalization in Fig.~\ref{fig:view_gen}, liquid generalization in Fig.~\ref{fig:pull} and Table~\ref{tab:liquids}, objects–container–ladle generalization in Fig.~\ref{fig:pull}, Table~\ref{tab:real1} and Table~\ref{tab:level2}, and robustness to severe occlusions in Fig.~\ref{fig:meatball_scoop}, Table~\ref{tab:simulation_six_objs_occlusion_unify} and Table~\ref{tab:real_occlusion}.
Moreover, our approach can be directly transferred to other robotic platforms and rapidly deployed in new environments.
While \method is evaluated on the scooping task, the underlying principles—such as object-centric perception, efficient data collection strategy, and coarse-to-fine policy design can also inspire solutions for other manipulation tasks involving fluid–solid interaction, or dynamic object handling.


\begin{thebibliography}{54}


\bibitem{sundaresan2023acquisition}Sundaresan et al. Learning Sequential Acquisition Policies for Robot-Assisted Feeding. {\em CoRL}. (2023)
\bibitem{liu2024grounding}Liu et al. Grounding DINO: Marrying DINO with Grounded Pre-training for Open-set Object Detection. {\em ECCV}. (2024)
\bibitem{tai2023scone}Tai et al. SCONE: A Food Scooping Robot Learning Framework with Active Perception. {\em CoRL}. (2023)
\bibitem{coumans2019}Coumans, E. \& Bai, Y. PyBullet, a Python Module for Physics Simulation for Games, Robotics and Machine Learning. (http://pybullet.org)
\bibitem{Seita2022toolflownet}Seita et al. ToolFlowNet: Robotic Manipulation with Tools via Predicting Tool Flow from Point Clouds. {\em CoRL}. (2022)
\bibitem{oquab2024dinov}Oquab et al. DINOv2: Learning Robust Visual Features without Supervision. {\em Transactions On Machine Learning Research}. (2024)
\bibitem{corl2020softgym}Lin et al. SoftGym: Benchmarking Deep Reinforcement Learning for Deformable Object Manipulation. {\em CoRL}. (2020)
\bibitem{cad2rl}Sadeghi, F. \& Levine, S. CAD2RL: Real Single-Image Flight without a Single Real Image. {\em RSS}. (2017)
\bibitem{schenck2017learning}Schenck et al. Learning Robotic Manipulation of Granular Media. {\em CoRL}. (2017)
\bibitem{ziebart2008irl}Ziebart et al. Maximum Entropy Inverse Reinforcement Learning. {\em AAAI}. (2008)
\bibitem{DiffusionPolicy2023}Chi et al. Diffusion Policy: Visuomotor Policy Learning via Action Diffusion. {\em RSS}. (2023)
\bibitem{liu2022robotcookingstirfry}Liu et al. Robot Cooking With Stir-Fry: Bimanual Non-Prehensile Manipulation of Semi-Fluid Objects. {\em RA-L}. (2022)
\bibitem{reality_gap_1995}Jakobi et al. Noise and the Reality Gap: The use of Simulation in Evolutionary Robotics. {\em Advances In Artificial Life}. (1995)
\bibitem{jenamani2024flair}Jenamani et al. FLAIR: Feeding via Long-horizon AcquIsition of Realistic dishes. {\em RSS}. (2024)
\bibitem{PoseCNN}Xiang et al. PoseCNN: A Convolutional Neural Network for 6D Object Pose Estimation in Cluttered Scenes. {\em RSS}. (2018)
\bibitem{narasimhan2022pouring}Narasimhan et al. Self-supervised Transparent Liquid Segmentation for Robotic Pouring. {\em ICRA}. (2022)
\bibitem{PointNet2_2017}Qi et al. PointNet++: Deep Hierarchical Feature Learning on Point Sets in a Metric Space. {\em NeurIPS}. (2017)
\bibitem{bhaskar2024lava}Bhaskar et al. LAVA: Long-horizon Visual Action based Food Acquisition. {\em IROS}. (2024)
\bibitem{calli2017ycbdata}Calli et al. Yale-CMU-Berkeley Dataset for Robotic Manipulation Research. {\em IJRR}. (2017)
\bibitem{diff_fluid_dynamics_2018}Schenck et al. SPNets: Differentiable Fluid Dynamics for Deep Neural Networks. {\em CoRL}. (2018)
\bibitem{schenck2017visualclosedloopcontrolpouring}Schenck et al. Visual Closed-Loop Control for Pouring Liquids. {\em ICRA}. (2017)
\bibitem{mujoco}Todorov et al. MuJoCo: A Physics Engine for Model-Based Control. {\em IROS}. (2012)
\bibitem{openai-dactyl}OpenAI, Andrychowicz et al. Learning Dexterous In-Hand Manipulation. {\em IJRR}. (2019)
\bibitem{deformables_survey_2022}Zhu et al. Challenges and Outlook in Robotic Manipulation of Deformable Objects. {\em ArXiv:2105.01767}. (2021)
\bibitem{lin2023pourit}Lin et al. PourIt!: Weakly-supervised Liquid Perception from a Single Image for Visual Closed-Loop Robotic Pouring. {\em ICCV}. (2023)
\bibitem{ALOHA}Zhao et al. Learning Fine-Grained Bimanual Manipulation with Low-Cost Hardware. {\em RSS}. (2023)
\bibitem{rubik_cube_2019}OpenAI, Akkaya et al. Solving Rubik's Cube with a Robot Hand. {\em ArXiv:1910.07113}. (2019)
\bibitem{ho2020ddpm}Ho et al. Denoising Diffusion Probabilistic Models. {\em NeurIPS}. (2020)
\bibitem{domain_randomization}Tobin et al. Domain Randomization for Transferring Deep Neural Networks from Simulation to the Real World. {\em IROS}. (2017)
\bibitem{qi2024learninggeneralizabletooluseskills}Qi et al. Learning Generalizable Tool-use Skills through Trajectory Generation. {\em IROS}. (2024)
\bibitem{pomerleau1989alvinn}Pomerleau, D. Alvinn: An Autonomous Land Vehicle in a Neural Network. (Carnegie-Mellon University,1989)
\bibitem{xian2023fluidlab}Xian et al. FluidLab: A Differentiable Environment for Benchmarking Complex Fluid Manipulation. {\em ICLR}. (2023)
\bibitem{imitation_survey_2018}Osa et al. An Algorithmic Perspective on Imitation Learning. {\em Foundations And Trends In Robotics}. \textbf{7} (2018)
\bibitem{bimanual_food_2022}Grannen et al. Learning Bimanual Scooping Policies for Food Acquisition. {\em CoRL}. (2022)
\bibitem{niu2023goatsgoalsampling}Niu et al. GOATS: Goal Sampling Adaptation for Scooping with Curriculum Reinforcement Learning. {\em IROS}. (2023)
\bibitem{manip_deformable_survey_2018}Sanchez et al. Robotic Manipulation and Sensing of Deformable Objects in Domestic and Industrial Applications: a Survey. {\em IJRR}. (2018)
\bibitem{li2022behavior}Li et al. BEHAVIOR-1K: A Benchmark for Embodied AI with 1,000 Everyday Activities and Realistic Simulation. {\em CoRL}. (2022)
\bibitem{daxbench_2023}Chen et al. DaXBench: Benchmarking Deformable Object Manipulation with Differentiable Physics. {\em ICLR}. (2023)
\bibitem{ravi2024sam2}Ravi et al. SAM 2: Segment Anything in Images and Videos. {\em ArXiv Preprint}. (2024)
\bibitem{makoviychuk2021isaac}Makoviychuk et al. Isaac Gym: High Performance GPU-Based Physics Simulation For Robot Learning. {\em ArXiv:2108.10470}. (2021)
\bibitem{ruangpayoongsak2017wastescooper}Ruangpayoongsak et al. A Floating Waste Scooper Robot On Water Surface. {\em ICCAS}. (2017)
\bibitem{brose2010assistive}Brose et al. The Role of Assistive Robotics in the Lives of Persons with Disability. {\em American Journal of Physical Medicine \& Rehab}. (2010)
\bibitem{seita_fabrics_2020}Seita et al. Deep Imitation Learning of Sequential Fabric Smoothing From an Algorithmic Supervisor. {\em IROS}. (2020)
\bibitem{park2017multimodal}Park et al. A Multimodal Execution Monitor with Anomaly Classification for Robot-assisted Feeding. {\em IROS}. (2017)
\bibitem{Bommasani2021FoundationModels}Bommasani et al. On the Opportunities and Risks of Foundation Models. {\em ArXiv}. (2021), https://crfm.stanford.edu/assets/report.pdf
\bibitem{pyrender}Matl, M. Pyrender. (2019), https://github.com/mmatl/pyrender
\bibitem{hsu2022visionbasedmanipulatorsneedhands}Hsu et al. Vision-Based Manipulators Need to Also See from Their Hands. {\em ICLR}. (2022)
\bibitem{Black20240AV}Black et al. $\pi_0$: A Vision-Language-Action Flow Model for General Robot Control. {\em ArXiv:2410.24164} (2024)
\bibitem{Lyu2024ScissorBotLG}Lyu et al. ScissorBot: Learning Generalizable Scissor Skill for Paper Cutting via Simulation, Imitation, and Sim2Real. {\em CoRL}. (2024)
\bibitem{Dasari2022LearningDM}Dasari et al. Learning Dexterous Manipulation from Exemplar Object Trajectories and Pre-Grasps. {\em ICRA}. (2022)

\end{thebibliography}
\end{document}